\documentclass{article}
\usepackage{arxiv}

\usepackage[utf8]{inputenc} 
\usepackage[T1]{fontenc}    
\usepackage{hyperref}       
\usepackage{url}            
\usepackage{booktabs}       
\usepackage{amsfonts}       
\usepackage{nicefrac}       
\usepackage{microtype}      

\usepackage{graphicx}
\usepackage{comment}
\usepackage{amsmath,amssymb} 
\usepackage{color}
\usepackage{algpseudocode}

\definecolor{red}{rgb}{1.00,0.00,0.00}
\definecolor{blue}{rgb}{0.00,0.00,1.00}
\definecolor{green}{rgb}{0.2,0.70,0.2}
\definecolor{yellow}{rgb}{0.5,0.5,0.0}

\usepackage{graphicx}
\usepackage{amsmath,amssymb,amsfonts}
\usepackage{hyperref}
\usepackage[ruled,vlined]{algorithm2e}
\usepackage{multirow}
\usepackage{tabularx, booktabs}
\usepackage{floatrow}
\usepackage{subfig}
\usepackage{mathtools}
\usepackage{arydshln}
\usepackage{xcolor}
\usepackage{mwe}
\usepackage{floatrow}%
\usepackage{wrapfig}
\usepackage[bottom]{footmisc} 
\usepackage{xcolor}
\newcommand{\myRed}[1]{\textcolor{black}{#1}}

\title{Local-HDP: Interactive Open-ended  3D Object \\ Category Recognition  in real-time robotic scenarios}

\author{%
 H. Ayoobi$^\star$,\;\; H. Kasaei,\;\;
        M. Cao,\;\;
        R. Verbrugge,\;\;
        B. Verheij \\
Bernoulli Institute, University of Groningen, Netherlands \\
$^\star$h.ayoobi@rug.nl
}

\begin{document}

\maketitle

\begin{abstract}
We introduce a non-parametric hierarchical Bayesian approach for open-ended 3D object categorization, named the Local Hierarchical Dirichlet Process (Local-HDP). This method allows an agent to learn independent topics for each category incrementally and to adapt to the environment in time. Hierarchical Bayesian approaches like Latent Dirichlet Allocation (LDA) can transform low-level features to high-level conceptual topics for 3D object categorization. However, the efficiency and accuracy of LDA-based approaches depend on the number of topics that is chosen manually. Moreover, fixing the number of topics for all categories can lead to overfitting or underfitting of the model. In contrast, the proposed Local-HDP can autonomously determine the number of topics for each category. Furthermore, the online variational inference method has been adapted for fast posterior approximation in the Local-HDP model. Experiments show that the proposed Local-HDP method outperforms other state-of-the-art approaches in terms of accuracy, scalability, and memory efficiency by a large margin. 
Moreover, two robotic experiments have been conducted to show the applicability of the proposed approach in real-time applications.
\end{abstract}
\vspace{2mm}
\section{Introduction}

\begin{figure}[H]

\begin{tabular}{c c c }
  \includegraphics[width=0.35\linewidth]{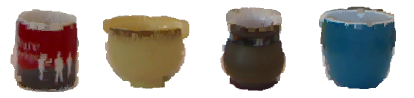} &\;\;\;\;\;\; &
  \includegraphics[width=0.35\linewidth]{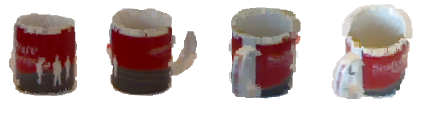} 
\end{tabular}
\caption{ An illustrative example of (\textit{left})  intra-category variation of the mug category in the Washington RGB-D dataset, and (\textit{right}) different object views of a mug object. }
\label{FigMugsVarView}
\end{figure}

Most recent object recognition/detection techniques are based on deep neural networks \cite{huang2019perspectivenet, 2019Kanezaki,2018Liang,2017Ren,2019Sangineto,2019Loghmani}. These methods typically need a large labeled dataset for a long training process. The number of object categories (class labels) should be predefined in advance for such methods. However, in real-life robotic scenarios, a robot can always face new object categories while operating in its environment. Therefore, the model should get updated in an open-ended manner without completely retraining the model \cite{hamed2019}.
Furthermore, object category recognition is not a well-defined problem because of the large inter-category variation (Figure~\ref{FigMugsVarView}~(\textit{left})), multiple object views for each object (Figure~\ref{FigMugsVarView}~(\textit{right})), and concept drift in dynamic environments \cite{kasaei2019local}.

Object recognition in humans is a complex hierarchical multi-stage process of streaming visual data in the cortical regions \cite{cichy2016comparison}. The hierarchical structure of the brain for the object recognition task has motivated us to choose hierarchical Bayesian models like Latent Dirichlet Allocation (LDA) \cite{blei2003latent} and Hierarchical Dirichlet Process (HDP) \cite{Teh2006HDP2} for object category recognition. 

In this paper, we suggest that 3D visual streaming data should be processed continuously, and object category learning and recognition should be performed simultaneously in an open-ended manner.We propose the Local Hierarchical Dirichlet Process (Local-HDP), an extension of the Hierarchical Dirichlet Process \cite{Teh2006HDP2} method, which can incrementally learn new topics for each category of objects independently. 
In contrast to notable recent works \cite{kasaei2019local,IROS,shen2017topic} using a predefined number of topics, Local-HDP is more flexible since it is a non-parametric Bayesian model that can autonomously determine the number of topics for each category at run-time.

\setlength{\columnsep}{14pt}%
\setlength{\intextsep}{11pt}%
\begin{wrapfigure}{r}{0.5\textwidth}
\vspace{-5mm}
  \begin{center}
    \includegraphics[width=1\linewidth]{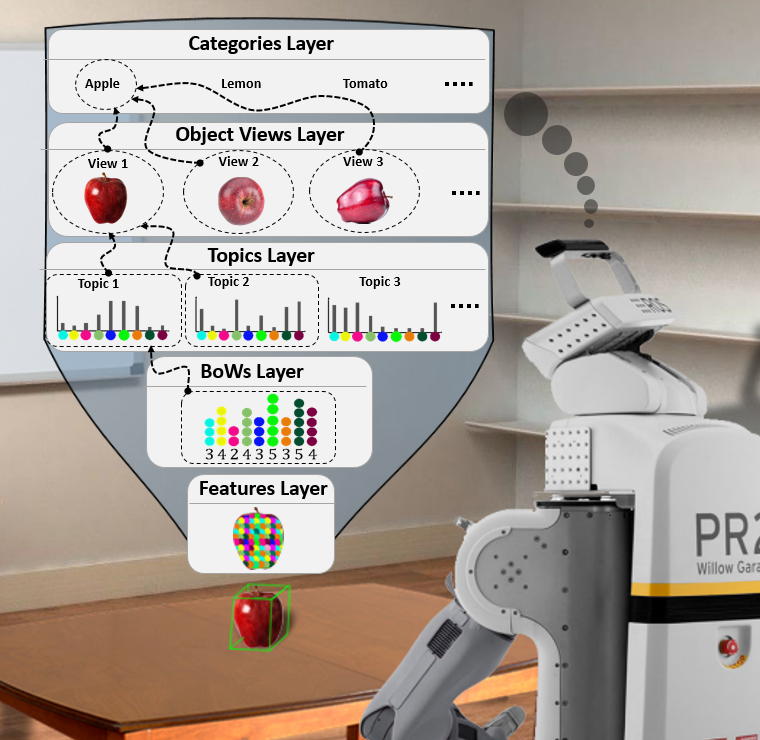}
  \end{center}
   \caption{The architecture of the proposed method. }
   
\label{fig:arc}
\end{wrapfigure}

Figure~\ref{fig:arc} shows the processing layers of the proposed Local-HDP. The tabletop objects are detected in the initial phase (green bounding box around apple on the table in Figure~\ref{fig:arc}). Subsequently, the hierarchy of the five processing layers is utilized. The features layer extracts a set of local shape features using the spin-image descriptor~\cite{johnson1999using}. 
The computed features are represented as Bag of visual Words (BoWs). The obtained representation is then sent to the topics layer, where a set of topics is inferred autonomously for the given object using the proposed Local-HDP method. Each topic is a distribution over visual words. In other words, the topic layer provides an unsupervised mapping of the BoW representation to the topics space, which can fill the conceptual gap between low-level features and high-level concepts. As shown in the object views layer, the appearance of an object may vary from different perspectives~(Figure~\ref{FigMugsVarView}~(\textit{bottom})). Therefore, it is necessary to infer topics using different object views. There might be different instances in an object category as well (see Figure~\ref{FigMugsVarView}~(\textit{top})). This point is addressed in the categories layer. Moreover, a simulated teacher has been developed to interact with the model and evaluate its performance in an open-ended manner.

This work extends two approaches, namely Local-LDA \cite{kasaei2019local} and HDP \cite{Teh2006HDP2}, in four aspects. First, our approach can autonomously detect the number of required topics to independently represent the objects in each category, avoiding the limitation of Local-LDA for determining the number of topics in advance. This feature prevents underfitting or overfitting of the model. Second, our research adapts the online variational inference technique \cite{pmlr-v15-wang11a}, which significantly reduces inference time. Third, the proposed local online variational inference method leads to memory optimization since it needs to store a smaller average number of instances per object category in memory.  Fourth, our work extends the hierarchical Dirichlet process \cite{Teh2006HDP2} by learning and updating local topics for each object category independently in an incremental and open-ended fashion. 
\section{Related Work}
\label{related_work}
Object representation is one of the main building blocks of object recognition approaches. The underlying reason is that the output of the object representation module is used in both learning and recognition. 
Object representation techniques can be categorized into three groups, namely, global and local object descriptors and machine learning approaches \cite{laga20183d}. Notable global object descriptors are Global Orthographic Object Descriptor (GOOD)~\cite{GOOD,kasaei2016orthographic}, Ensemble of Shape Functions (ESF)~\cite{ESF} and Viewpoint Feature Histogram (VFH) \cite{VFH}.  Examples of local 3D shape descriptors include Spin-Images (SI) \cite{johnson1999using}, Intrinsic Shape Signature (ISS) \cite{zhong2009intrinsic}, and Fast Point Feature Histogram (FPFH) \cite{rusu2009fast}.  Local descriptors are more robust to occlusions and clutter. However, comparing pure local descriptors is a computationally expensive task \cite{aldoma2012tutorial}. 
To alleviate this problem, machine learning techniques like Bag of Words (BoW) \cite{kasaei2015adaptive}, Latent Dirichlet Allocation (LDA) \cite{blei2003latent,kasaei2016hierarchical} and deep learning \cite{li2016fpnn,wu20153d} methods can be used for representing objects in a compact and uniform format.

Kasaei et al. \cite{kasaei2019local} extended Latent Dirichlet Allocation (LDA) and proposed Local-LDA. They showed the application of Local-LDA in the context of open-ended 3D object category learning and recognition. Similar to our approach, Local-LDA learns a set of topics for each object category incrementally and independently. Unlike our approach, in Local-LDA, the same number of topics is chosen in advance based on trials and errors for all of the object categories. A good choice for the number of topics for each object category is correlated to the intra-category variation of each 3D object category. Therefore, choosing the same number of topics for all the object categories with different intra-category variation might be not reasonable. Moreover, in open-ended scenarios, it is not feasible to anticipate the inter-category variation of 3D objects that the model might see in the future and choose a fixed number of topics in advance for all the categories. To solve these issues, our approach can autonomously choose the number of topics for each object category on the fly without a need for in advanced trails and errors. This makes our approach more robust for recognizing object categories with various inter-category and intra-category variation and applicable in real-world open-ended scenarios.  Local-LDA uses collapsed Gibbs sampling for approximating the posterior probability. However, we adapt the online variational inference technique \cite{pmlr-v15-wang11a} for Local-HDP.

Our approach builds on the Hierarchical Dirichlet Process (HDP) \cite{Teh2006HDP2}, that is based on Dirichlet process (DP) \cite{ferguson1973bayesian} and mixture of DPs \cite{antoniak1974mixtures}. Posterior inference is intractable for HDP, and much research has been done to find a proper approximate inference algorithm \cite{Teh2006HDP2,teh2008collapsed,liang2007infinite}. The Markov Chain Monte Carlo (MCMC) sampling method for DP mixture models has been proposed for approximate inference in HDPs \cite{neal2000markov}. David Blei et al. proposed the variational inference for DP mixtures \cite{blei2006variational}. Teh et al. \cite{Teh2006HDP2} proposed the Chinese Restaurant Franchise metaphor for HDP and used Gibbs sampling method for the inference. The online variational inference approach is proposed by Wang et al. \cite{pmlr-v15-wang11a} for HDP, which can be used in online incremental learning scenarios and for large corpora. Our method is different from HDP, since HDP only shares the topics among the same categories and not across different categories. This is especially needed in the case of 3D object categorization for open-ended scenarios \cite{kasaei2019local}. HDP has further extensions to construct tree-structured representations for text data which have nested structure \cite{Blei2015}. Similar to supervised hierarchical Dirichlet Process (sHDP) \cite{Dai2015}, we use the category label of each object.  Unlike sHDP, we learn object categories in an open-ended fashion, while in sHDP, the number of object categories to be learned should be defined in advance.

 Deep learning-based approaches \cite{zhou2018voxelnet,Yu_2018_CVPR,Feng_2018_CVPR} try to learn a sparse representation for 3D objects. Unlike our approach, such methods typically need a large labeled dataset and require long training time. In particular, our proposed approach does not require a large labeled dataset and can incrementally update the model facing an unforeseen object category in an open-ended manner. Moreover, the number of categories is not fixed in open-ended approaches like ours.  
 
\section{Method}
 \label{method}
 We assume that an object has already been segmented from the point cloud of the scene, and we hence mainly focus on detailing the Local Hierarchical Dirichlet Process (Local-HDP) approach. 

\subsection{Pre-Processing Layers} 

\begin{figure}[!t]
\centering
\begin{tabular}{c c c c c c c}
  \includegraphics[width=0.12\linewidth]{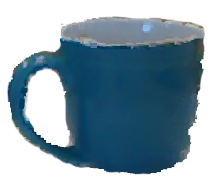} 
  &\;\; &
  \includegraphics[width=0.12\linewidth]{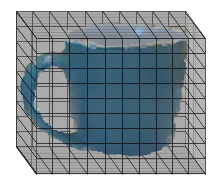} 
  &\;\; &
  \includegraphics[width=0.12\linewidth]{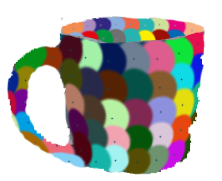}
  &\;\; &
  \includegraphics[width=0.12\linewidth]{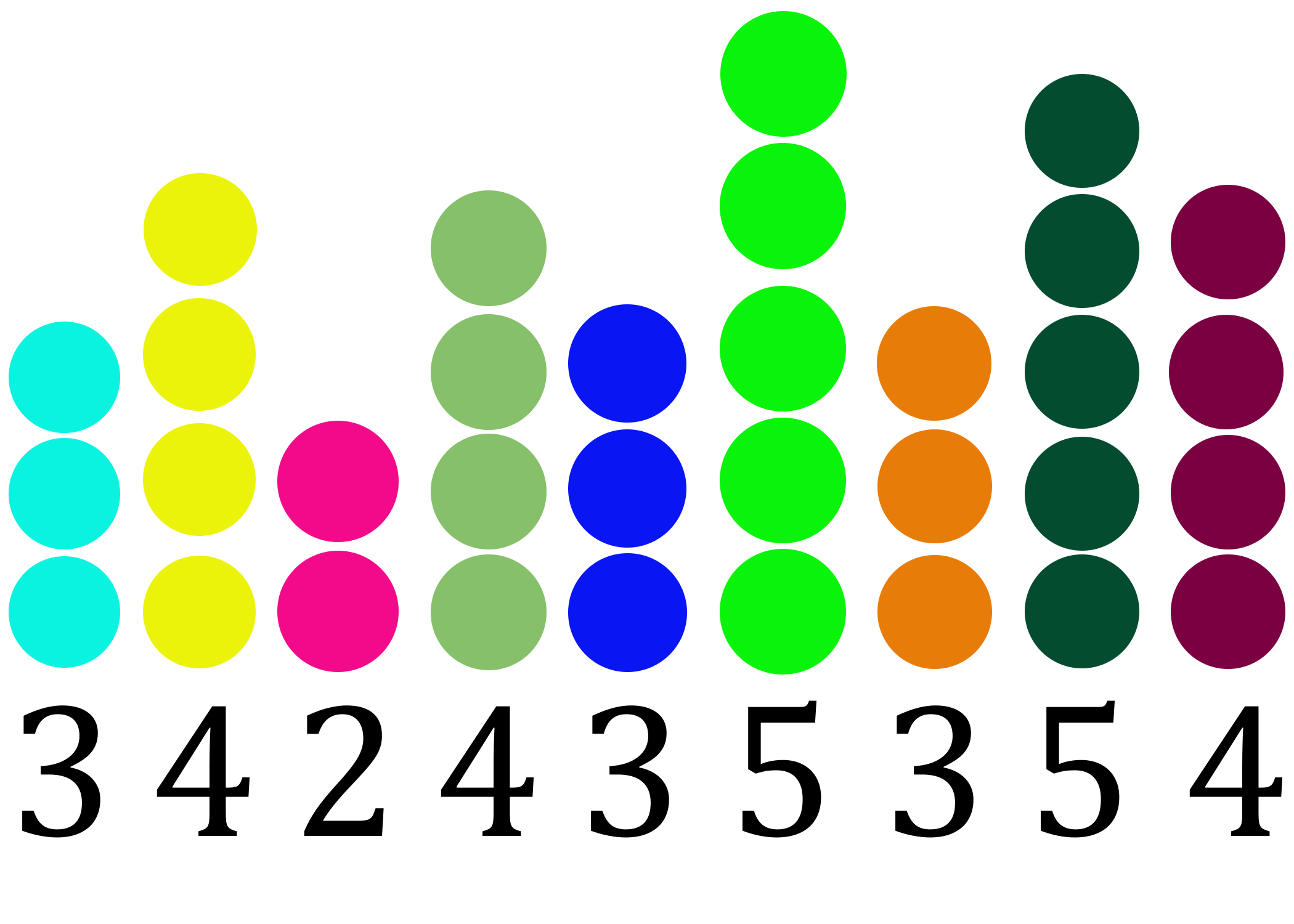}\\
(\textit{a}) coffee mug  &\;\;& (\textit{b}) voxelization &\;\;&
 (\textit{c}) local-features  &\;\;& (\textit{d}) BoW\\ 
\end{tabular}
\caption{ (\textit{a}) The RGB-D image of a coffee mug. (\textit{b}) Key-points selection using voxelizing \cite{kasaei2019local}. (\textit{c}) Key-points neighborhoods are represented by different colors.  (\textit{d}) The BoW representation for the given object.}
\label{FigMug}
\end{figure}

In Figure \ref{fig:arc}, the first two layers---the feature layer and BoWs layer--- are the pre-processing layers. In the feature layer, we first select key-points for the given object and then compute a local shape feature for each key-point. Towards this goal, we first voxelized\footnote{\url{http://docs.pointclouds.org/trunk/classpcl_1_1_voxel_grid.html}} the object (Figure~\ref{FigMug}) (\textit{b}), and then, the nearest point to each voxel center is selected as a key-point. Afterwards, the spin-image descriptor~\cite{johnson1999using} is used to encode the surrounding shape in each key-point using the original point cloud (Figure~\ref{FigMug} (\textit{c})). This way, each object view is described by a set of spin-images in the first layer, $\mathbf{O_s} = \{\mathbf{s}_1,\dots, \mathbf{s}_N\}$ where $N$ is the number of key-points. The obtained representation is then sent to the BoWs layer. Since HDP-based models have the bag-of-words assumption - that the order of words in the document can be neglected - the BoWs layer transforms the computed spin-images to a BoW format (Figure~\ref{FigMug} (\textit{d})). Towards this end, the BoWs layer requires a dictionary with $V$ visual words (spin-images). In this work, we have created a dictionary of visual words using the same methodology as Local-LDA \cite{kasaei2019local}. The obtained BoW representation is fed to the topic layer.

\subsection{Local Hierarchical Dirichlet Process}
After synthesizing the point cloud of the 3D objects to a set of visual words in BoW format, the data is ready to be inserted into the topic layer where the proposed Local-HDP method is employed. In this layer, the model transforms the low-level features in BoW format to conceptual high-level topics. In other words, each object is represented as a distribution over topics, where each topic is a distribution over visual words. To this end, we use an incremental inference approach where the number of categories is not known beforehand and the agent does not know which additional object categories will be available at run-time. The plate notation of Local-HDP is shown in Figure \ref{figLocHDP}. In this graph, $C$ is the number of categories, $|c|$ is the number of objects in each category. Each object, $d$, is represented by a set of $N$ visual words, $W_{d,n}$ where $n,d$ show the n'th visual word from the d'th object. Each visual word is an element from the vocabulary of visual worlds with predefined $V$ words, that is $W_{d,n} \in \{1...V\}$. Using a \textit{Coffee Mug} as an example, a distribution over the topics of the \textit{Coffee Mug} should be used to generate the visual words of the object. Accordingly, a particular topic is selected out of the mixture of possible topics of the \textit{Coffee Mug} category to generate the visual words. For instance, coffee mugs typically have a ``handle'', which is represented as a distribution of visual words that repeatedly occurring together. This can be interpreted as the ``handle'' topic, which is inferred from the co-occurrence of the visual words in several objects of the same category. The process of choosing a topic and then drawing the visual words from that topic is repeated several times to generate all the visual words of the \textit{Coffee Mug}.
After constructing the model in a generative manner, a reverse procedure for inferring the latent variables from the data is used.

\subsection{Local Online Variational Inference}
In this section, we adapt the online variational inference approximate inference method  \cite{pmlr-v15-wang11a} for Local-HDP. This method can be used in open-ended applications since it can handle streaming data in an online and incremental manner. Moreover, it is faster than traditional approximate inference techniques, e.g., Chinese restaurant franchise \cite{Teh2006HDP2} and variational inference \cite{blei2006variational}, and it can be used to infer the latent variables of different scale datasets~\cite{pmlr-v15-wang11a}.

Online variational inference for HDP is inspired by the online variational Bayes \cite{hoffman2010online} method for LDA. This method tries to optimize a variational objective function \cite{jordan1999introduction} exploiting stochastic optimization \cite{robbins1951stochastic}. Using Sethuraman’s stick-breaking construction for HDP \cite{Teh2006HDP2}, the variational distribution for local online variational inference is in the following form:
\vspace{-1mm}
\begin{equation}
    q(\beta', \pi', c, z, \phi) = q(\beta')q(\pi')q(c)q(z)q(\phi)
\end{equation}

\myRed{In the terminology of variational inference techniques, \textit{q} is called the variational approximation to the posterior \textit{p}. Variational techniques try to solve an optimization problem over a class of tractable distributions \textit{Q} in order to find a $q \in Q$ that is most similar to \textit{p} and can be used as its approximation}. Moreover, $\beta'=(\beta_{k}^{'})_{k=1}^{\infty}$ is the top-level stick proportion, $\pi'=(\phi_{jt}^{'})_{t=1}^{\infty}$ is the bottom-level stick proportion and $c_j = (c_{jt}^{'})_{t=1}^{\infty}$ is the vector of indicators for each $G_j$. Moreover, $\phi = (\phi_k)_{k=1}^{\infty}$ is the inferred topic distribution, and $z_{jn}$ is the topic index for \myRed{the \textit{n}th word in the \textit{j}th document $w_{jn}$.}

The factorized form of  $q(c),\; q(z),\; q(\phi) ,\; q(\beta')\; \text{and}\; q(\pi')$ is the same as the online variational inference for HDP \cite{wang2011online}. Assuming that we have $|c|$ objects in each category for Local-HDP, the variational lower bound for object $j$ in category $C$ is calculated as follows:
\vspace{-1mm}

\begin{center}
\begin{multline}
    L_j^{(C)}=  \mathop{{}\mathbb{E}_q}[log(p(w_j|c_j,z_j,\phi)p(c_j|\beta')p(z_j|\pi')p(\pi_j^{'}|\alpha_0))] +   H(q(c_j))  + H(q(z_j)) \\ + H(q(\phi')) +   \frac{1}{|c|}[E_q[log p(\beta')p(\phi)] + H(q(\beta')) + H(q(\phi)]
\end{multline}
\end{center}

Where $H(.)$ is the entropy term for the variational distribution. Therefore, the lower bound term for each category is calculated in the following way:

\begin{equation}
    L^{(C)} = \sum_{j} L_j^{(C)} = \mathop{{}\mathbb{E}_j}[|c|L_j^{(C)}]
\end{equation}

Using coordinate ascent equations in the same way as online variation inference, the object-level parameters $(a_j, b_j, \myRed{\varphi_j}, \zeta_j)$ are estimated. \myRed{To be more specific, $a_j$ and $b_j$ are the parameters of the beta distributions for the bottom-level stick proportions $\pi_j$, $\varphi_j$ is the variational parameter for the vector of indicators $c_j$, and $\zeta_j$ is the variational parameter for the topic $z_j$. These variables are defined in the same way as in \cite{wang2011online}}. Then, for the category-level parameters $(\lambda^{(C)}, u^{(C)},v^{(C)})$, we do gradient descent with respect to a learning rate:

\vspace{-4mm}\begin{equation}
    \partial \lambda_{kw}^{(C)}(j) = -\lambda_{kw} + \eta + |c| \sum_{t=1}^{T} \varphi_{jtk} (\sum_n \zeta_{jnt}I[w_{jn} = w])
\end{equation}
\vspace{-1.5mm}\begin{equation}
    \partial u_{k}^{(C)}(j) = -u_{k} + 1 + |c| \sum_{t=1}^{T} \varphi_{jtk} 
\end{equation}
\vspace{-1.5mm}\begin{equation}
    \partial v_{k}^{(C)}(j) = -v_{k} + \lambda + |c| \sum_{t=1}^{T}  \sum_{l=k+1}^{K} \varphi_{jtl}
\end{equation}

Here, K and T are the document and corpus level truncates. Moreover,  $\varphi$ (multinomial), $\zeta$ (multinomial) and $\lambda$ (Dirichlet) are the variational parameters, which are the same for all the categories. Using an appropriate learning rate $p_{t_0}$ for online inference, the updates for $\lambda^{(C)}, u^{(C)}$ and $v^{(C)}$ become:

\vspace{-2mm}\begin{equation}
    \boldsymbol{\lambda}^{(C)} \leftarrow \boldsymbol{\lambda}^{(C)} + p_{t_0} \partial \boldsymbol{\lambda}^{(C)}(j)
\end{equation}
\begin{equation}
    \boldsymbol{u}^{(C)} \leftarrow \boldsymbol{u}^{(C)} + p_{t_0} \partial \boldsymbol{u}^{(C)}(j)
\end{equation}
\begin{equation}
    \boldsymbol{v}^{(C)} \leftarrow \boldsymbol{v}^{(C)} + p_{t_0} \partial \boldsymbol{v}^{(C)}(j)
\end{equation}

Algorithm \ref{algLocalHDP} shows the pseudo-code of the proposed inference technique for the Local-HDP approach.

\begin{minipage}{.34\textwidth}
  \centering
  \begin{tabular}{c}
  \includegraphics[width=0.92\linewidth, trim={0cm, -1cm, 0cm, 0cm, 0mm}]{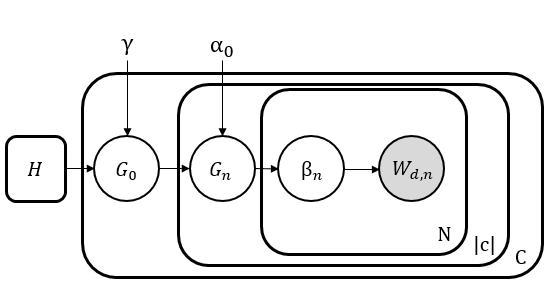}\vspace{-2mm}\\{\scriptsize (\textit{a}) Local-HDP\vspace{-2mm}}\\ \;\\
  \includegraphics[width=0.57\linewidth,height=2.5cm]{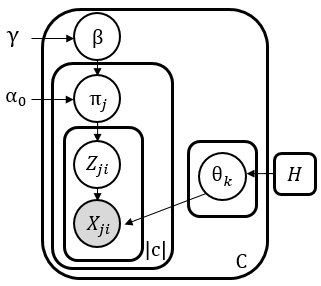}\\
  {\scriptsize(\textit{b}) Stick-breaking }
\end{tabular}
\captionof{figure}{The plate notation of Local-HDP and its stick-breaking construction.} \label{figLocHDP}
\end{minipage}
\begin{minipage}{.02\textwidth}
\;
\end{minipage}
\begin{minipage}{.61\textwidth}
\SetAlCapNameFnt{\small}
\SetAlCapFnt{\small}
\vspace{4mm}
\begin{algorithm}[H]
\small
\SetAlgoLined
\textbf{initialization:}\\
 Randomly initialize $\lambda^{(C)}= (\lambda_k^{(C)})_{k=1}^{K}$, $u^{(C)}= (u_k^{(C)})_{k=1}^{K-1}$ and $v^{(C)}= (v_k^{(C)})_{k=1}^{K-1}$ for all the learned categories. Set $t_0=1$\\ 
 \For{\textbf{each} Category C}{
     \While{Stopping criterion is not met}{
      - Use the object view j for updating the parameters.\\
      - Compute the document-level parameters $a_j, b_j, \Phi_j, \zeta_j$ using the same methodology as \cite{pmlr-v15-wang11a}.\\
      - Using Eq. 4-6, compute the natural gradients $\partial \lambda^{(C)}(j)$, $\partial u^{(C)}(j)$ and $\partial v^{(C)}(j)$.\\
      - Set $p_{t_0}=(\tau_0 + t_0)^{-K}$, $t_0 = t_0 + 1$.\\
      - Update the $\lambda^{(C)}$, $u^{(C)}$, $v^{(C)}$ parameters using Eq. 7-9.\\ 
     }
 }
 \normalsize
 \caption{Local Online Variational Inference}
 \label{algLocalHDP}
\end{algorithm}
\end{minipage}%
\vspace{4mm}

\subsection{Object Category Learning and Recognition}
In this subsection, the mechanism of interactive open-ended learning has been explained in more detail. Classical object recognition methods do not support open-ended learning. In contrast, our method is open-ended, and the number of categories can be incrementally extended through time. The system can interact with a human user to learn about new categories or to update existing category models by receiving corrective feedback when misclassification occurred. We follow the same methodology as \cite{Kasaei2015} for this purpose. The user can interact with the system with one of the following actions:

\begin{itemize}
    \item \textbf{Teach}: introducing the category of target object to the agent.
    \item \textbf{Ask}: inquiring the agent about the category of a target object.
    \item \textbf{Correct}: sending corrective feedback to the agent in case of wrong categorization. 
\end{itemize}

Whenever the agent receives a teach command, it incrementally updates the local model corresponding to the category of the target object using the aforementioned online variational inference technique. In case of the ask command, the log-likelihood is used to determine the category of an object. The log-likelihood is computed in the same way as in \cite{pmlr-v15-wang11a}. The local model with highest likelihood is then selected as the predicted category for an object.

\section{Experimental Results}
\label{results}

Following the same protocol as Local-LDA \cite{kasaei2019local} for interacting with a simulated teacher, two sets of experiments have been conducted to evaluate the performance of the proposed method.
For Local-HDP in all the experiments, we set $p_{t_0}=(\tau_0 + t_0)^{-K}$ where $K \in (0.5, 1]$ and $\tau_0 > 0$ as suggested by \cite{pmlr-v15-wang11a}.

\subsection {Datasets and Baselines for Comparison}
For offline evaluation of the proposed Local-HDP and the other state-of-the-art approaches, we have used the RGB-D restaurant object dataset \cite{Kasaei2015}. This dataset has 10 categories of objects and each category has a significant intra-category variation. It consists of 306 different object views for 10 household objects. Therefore, it is a suitable dataset to perform extensive sets of experiments. 

\begin{figure}
\setlength\arrayrulewidth{1pt}
\setlength\tabcolsep{0.5pt}
\begin{tabular}{c c c c c c c c c c c c}

   \includegraphics[width=.08\linewidth]{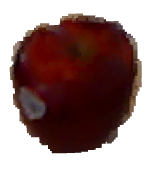}  &
    \includegraphics[width=.1\linewidth]{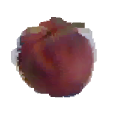}
    &
   \includegraphics[width=.09\linewidth]{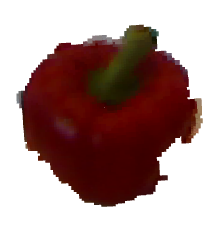}
   &
    \includegraphics[width=.1\linewidth]{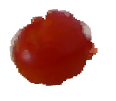}
   & 
    \includegraphics[width=.1\linewidth]{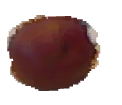}
    &
       \includegraphics[width=.1\linewidth]{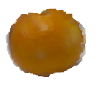}
   &
    \includegraphics[width=.09\linewidth]{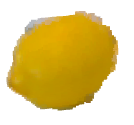}
   &
   \includegraphics[width=.09\linewidth]{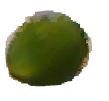}
   &
   \multicolumn{1}{:}{}
   &
    \includegraphics[width=.1\linewidth]{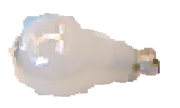}
   & 
    \includegraphics[width=.09\linewidth]{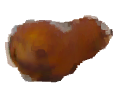}
    \\
   \includegraphics[width=.087\linewidth]{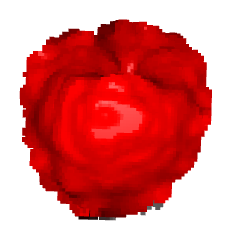}  &
    \includegraphics[width=.083\linewidth]{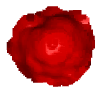}
    &
   \includegraphics[width=.092\linewidth]{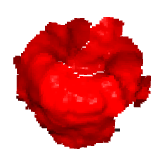}
   &
    \includegraphics[width=.087\linewidth]{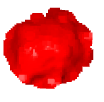}
   & 
    \includegraphics[width=.1\linewidth]{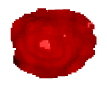}
    &
   \includegraphics[width=.1\linewidth]{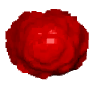}
   &
    \includegraphics[width=.1\linewidth]{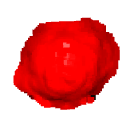}
    &
   \includegraphics[width=.1\linewidth]{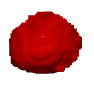}
   &
   \multicolumn{1}{:}{}
   &
    \includegraphics[width=.1\linewidth]{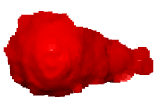}
   & 
    \includegraphics[width=.095\linewidth]{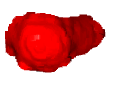}
\end{tabular}
\centering
\caption{RGB images for objects of different categories with depth data similarities in the Washington RGBD dataset.}
\label{figWashington}
\end{figure}

The Washington RGB-D dataset~\cite{WashingtonRGBDDS} is used for online open-ended evaluation of the method since it is one of the largest 3D object datasets. It has 250,000 views of 300 common household objects, categorized in 51 categories. Figure~\ref{figWashington} shows some of the categories of objects presented in the Washington RGBD Dataset. In all experiments, only the depth data has been used for determining the category of 3D objects. 
Therefore, as one can see in Figure~\ref{figWashington}, detecting the category of an object based solely on the depth data is a hard task even for humans.

We have compared the proposed Local-HDP using local online variational inference with Local-LDA~\cite{kasaei2019local}, LDA with shared topics~\cite{blei2003latent}, BoW~\cite{kasaei2015adaptive}, RACE ~\cite{oliveira20163d}, and HDP with shared topics and online variational inference~\cite{pmlr-v15-wang11a}.

\subsection{Offline Evaluation}
Similar to Local-LDA, our approach has several parameters that should be well selected to provide an appropriate balance between recognition performance, memory usage and computation time. In order to finetune the parameters of our proposed method for offline evaluation, 240 experiments have been conducted with different parameter values. The voxel grid approach has been used for down-sampling and finding the keypoints for the local descriptor. Voxel grid has Voxel Size (VS) parameter which determines the size of each voxel. Moreover, the spin-image local descriptor has two parameters, namely Image Width (IW)  and Support Length (SL).

\begin{wraptable}{r}{8.5cm}

\fontsize{8}{9}\selectfont
\newcolumntype{?}{!{\vrule width 0.2pt}}
\setlength\arrayrulewidth{0.8pt}
\setlength\tabcolsep{2pt} 

\begin{center}
\vspace{-2mm}
\begin{tabular}{ | c | c || c | c ? c | c | c ? c | c | c | c ? c | c| c | c | c | c |c | c | c | }
\hline
 \multicolumn{2}{|c||}{Parameters}&\multicolumn{2}{c?}{IW}&\multicolumn{3}{c?}{VS}&\multicolumn{4}{c?}{SL} \\
\hline
 \multicolumn{2}{|c||}{Value} & 4 & 8 & 0.01 &0.02 &0.03  &0.03 & 0.04 & 0.05 & 0.1  \\ 
\hline
\hline
\multirow{2}{*}{\shortstack{Average \\ accuracy (\%)}}&Local-LDA & \textbf{84} & 83 & 81 & 82 & \textbf{86} & 81 & 83 & 84  & \textbf{85}   \\ 
\cline{2-20}
&Local-HDP & \textbf{94} & 92 & 91 & 93 & \textbf{95} & 91 & 92 & 92 & \textbf{94}  \\ 
\hline
\end{tabular}
\end{center}

\fontsize{8}{9}\selectfont
\newcolumntype{?}{!{\vrule width 0.2pt}}
\setlength\arrayrulewidth{0.8pt}
\setlength\tabcolsep{2pt} 
\begin{tabular}{ | c | c || c | c ? c | c | c ? c | c | c | c ? c | c| c | c | c | c |c | c | c | c| }

\hline
 \multicolumn{2}{|c||}{Parameters}& \multicolumn{10}{c|}{Dictionary Size}\\
\hline
 \multicolumn{2}{|c||}{Value}& 40  & 50 & 60 & 70 & 80 & 90 & 100 & 200 & 500 & 2000 \\ 
\hline
\hline
\multirow{2}{*}{\shortstack{Average \\ accuracy (\%)}}&Local-LDA & 82 & 82 & 82 & 83& 85 & 85& 86 & 87 & 88& \textbf{90}  \\ 
\cline{2-20}
&Local-HDP & 91 & 92 & 92 & 92 & 92 & 93 & 93 & 94 & 95& \textbf{96} \\ 
\hline

\end{tabular}
\caption{Average accuracy of Local-HDP and Local-LDA based on 240 experiments with different parameter values.}
\label{tbl1}
\end{wraptable}

  In all experiments, the first level and second level concentration parameters are set to 1, chunk size for offline evaluation is set to 1, and the maximum number of topics is set to 100. All the other parameters are set to the default values as proposed in \cite{wang2011online} . Moreover, in all the experiments the LDA parameters are set to be the same values as described in \cite{kasaei2019local}. Since online variational inference is a stochastic inference technique, for each experiment the order of the data instances has been permuted 10 times and for each permutation 10-fold cross-validation has been used. Accordingly, the results have been averaged.

Table \ref{tbl1} shows the comparison of Local-HDP and Local-LDA with different parameter values. As one can see in this table, the proposed Local-HDP method outperforms Local-LDA which is the best among the other methods (see \cite{kasaei2019local}). Using the best parameter values based on Table \ref{tbl1} and the corresponding tables in \cite{kasaei2019local}, the accuracy of all the approaches is shown in Table \ref{tbl2}.

\setlength{\columnsep}{20pt}%
\setlength{\intextsep}{5pt}%
\begin{wraptable}{r}{8.5cm}
\vspace{-1mm}
\centering
\newcolumntype{?}{!{\vrule width 0.5pt}}
\setlength\arrayrulewidth{0.8pt}
\setlength\tabcolsep{1pt} 
\renewcommand{\arraystretch}{1}
\fontsize{8.5}{9.5}\selectfont
\begin{tabular}{ | c | c | c|}
\hline
 Approach & Accuracy (\%) & Run-time (s)  \\
\hline
 RACE \cite{oliveira20163d} & 87.0 & 1757.20  \\ 
\hline
BoW \cite{kasaei2015adaptive} & 89.0 &  \textbf{195.60} \\ 
\hline
LDA (shared topics) \cite{blei2003latent} & 88.0 & 227 \\ 
\hline
Local-LDA \cite{kasaei2019local} & 91.0 & 348 \\ 
\hline
HDP (shared topics) \cite{pmlr-v15-wang11a} & 90.33 & 233 \\ 
\hline
Local-HDP (our approach) & \textbf{97.11} &  352 \\ 
\hline
\end{tabular}
\caption{The comparison of different approaches using the best parameter values.\vspace{-11mm}}
\vspace{-2mm}
\label{tbl2}
\end{wraptable}

Table \ref{tbl2} shows that Local-HDP outperforms the other state-of-the-art methods in terms of accuracy with a large margin. In particular, the accuracy of Local-HDP was 97.11\%, which is around 6.11 percentage point (p.p.) better than Local-LDA, and 6.78, 9.11, 8.11, 10.11 p.p better than HDP, LDA, BoW and RACE approaches respectively. Moreover, Local-HDP has almost the same run-time as Local-LDA. 

\subsection{Open-Ended Evaluation}
In order to evaluate our model in an open-ended learning scenario, we used the Washington RGBD dataset~\cite{WashingtonRGBDDS}, and we have followed the same methodology as discussed in~\cite{kasaei2019local}. In particular, we have developed a simulated teacher which can interact with the model by either \textit{teaching} a new category to it or \textit{asking} the model to categorize the unforeseen object view. In case of wrong categorization of an object by the model, a \textit{correcting feedback} is sent to the model by the simulated teacher. In order to teach a new category, the simulated teacher presents three randomly selected object views of the corresponding category to the model. After teaching a new category, all of the previously learned categories are tested using a set of randomly selected unforeseen object views. Subsequently, the accuracy of category prediction is computed. In order to calculate the accuracy of the model at each point, a sliding window of size $3n$ is used, where $n$ is the number of learned categories. If the corresponding accuracy is higher than a certain threshold $\tau=0.66$ (which means that the number of true-positives is at least twice the number of wrong predictions), a new category will be taught by the simulated teacher to the model. If the learning accuracy does not exceed the threshold $\tau$ after a certain number of iterations ($100$ for our experiments), the teacher infers that the agent is not able to learn more categories and the experiment stops. More details on the online evaluation protocol which has been used in our experiments can be found in~\cite{IROS}.

Since the performance of open-ended evaluation may depend on the order of introducing categories and object views (randomly selected at the beginning of each experiment), 10 independent experiments have been carried out for each approach. Several performance measures have been used to evaluate the open-ended learning capabilities of the methods, namely: (\textit{i}) the number of Learned Categories ($\#$LC); (\textit{ii}) the number of Question/Correction Iterations ($\#$QCI) by the simulated user; (\textit{iii}) the Average number of stored Instances per Category (AIC) ; (\textit{iv}) Global Categorization Accuracy (GCA), which represents the overall accuracy in each experiment. These performance measures have the following interpretations. $\#$LC shows the open-ended learning capability of the model, which answers the following question: How capable is the model in learning new categories? $\#$QCI shows the length of the experiment (iterations). AIC represents the memory efficiency of the method. A lower average number of stored instances per category means a higher memory efficiency of the method. AIC is also related to the learning speed. A smaller AIC means that the method requires less observations to correctly recognize each category. $\#$GCA shows the accuracy of the model in predicting the right category for each object.

\begin{figure*}[h!]
\begin{center}
\newcolumntype{?}{!{\vrule width 1pt}}
\setlength\arrayrulewidth{0.5pt}
\setlength\tabcolsep{1pt} 
\renewcommand{\arraystretch}{1}
\setlength\tabcolsep{0.1pt}
\vspace{-3mm}
\centering
\begin{tabular}{c}
\hspace{-7mm}
\subfloat[Summary of experiments for LDA \vspace{-1mm}]{
\setlength\tabcolsep{0.1pt}
\begin{tabular}{c c c}
\setlength\tabcolsep{0.5pt}
\fontsize{8}{9}\selectfont
\begin{tabular}{ | c | c |c | c |c |}
\hline
 Exp\# & \#QCI & \#LC & AIC & GCA(\%)\\
\hline
1& 201	&	8	&	14.88	&	52.74  \\ 

2& 231	&	8	&	16.38	&	53.68  \\ 

3& 336	&	10	&	17.2	&	57.74  \\ 

4& 495	&	\textbf{15}	&	15.47	&	\textbf{62.22}  \\ 

5& 193	&	9	&	14.44	&	46.63  \\ 

6& 138	&	5	&	17.4	&	47.83  \\ 

7& 264	&	7	&	20.29	&	54.17  \\ 

8& 348	&	10	&	19.3	&	53.16  \\ 

9& 206	&	9	&	15.22   &   46.60  \\ 

10& 279	&	10	&	16.9	&	50.18  \\ 
\hline
\textbf{Avg.} & 269 &\textbf{9.1} & \textbf{16.74} & \textbf{51}   \\ 
\hline
\end{tabular}
 & 
 \hspace{-1mm}
 \begin{minipage}{.38\textwidth}
     \centering
     \includegraphics[width=1\linewidth,height=4cm]{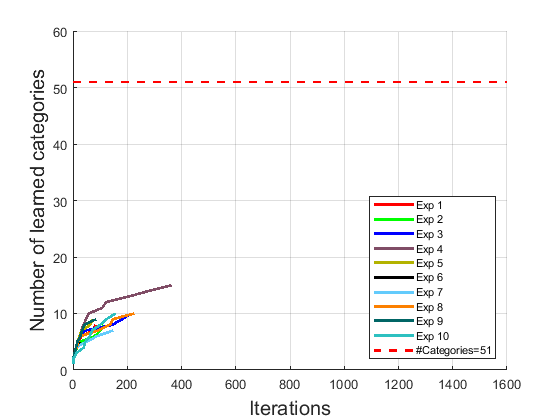}
\end{minipage}
 &
\hspace{-4mm}
 \begin{minipage}{.35\textwidth}
     \centering
     \includegraphics[width=1\linewidth,height=4cm]{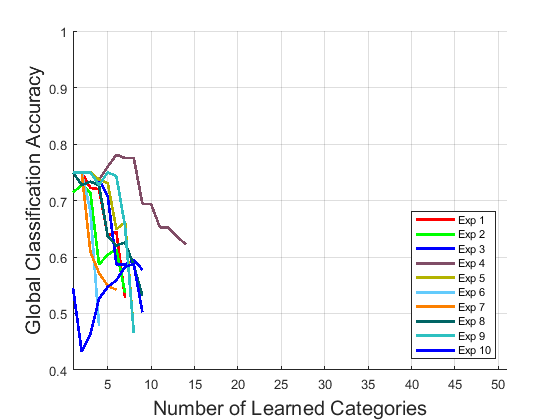}
\end{minipage}
\end{tabular}
}
\vspace{-3mm}
\\
\hspace{-7mm}
\subfloat[Summary of experiments for HDP \vspace{-1mm}]{
\setlength\tabcolsep{0.1pt}
\begin{tabular}{c c c}
\setlength\tabcolsep{0.5pt}
\fontsize{8}{9}\selectfont
\begin{tabular}{ | c | c |c | c |c |}
\hline
 Exp\# & \#QCI & \#LC & AIC & GCA(\%)\\
\hline
1& 1011	&	34	&	13.24	&	65.58  \\ 

2& 737	&	22	&	14.59	&	65.40  \\ 

3& 306	&	15	&	10.47	&	63.40  \\ 

4& 439	&	19	&	10.84	&	66.06  \\ 

5& 1079	&	34	&	13.26	&	67.66  \\ 

6& 1052	&	\textbf{35}	&	12.74	&	67.59  \\ 

7& 937	&	25	&	16.52	&	63.93  \\ 

8& 909	&	32	&	11.88	&	\textbf{68.76}  \\ 

9& 480	&	24	&	9.417	&	67.92  \\ 

10& 1069	&	32	&	14.66	&	65.11  \\ 
\hline
\textbf{Avg.} & 753 &\textbf{27.2} & \textbf{12.76} & \textbf{66.14}   \\ 
\hline
\end{tabular}
 & 
 \hspace{-1mm}
 \begin{minipage}{.37\textwidth}
     \centering
     \includegraphics[width=1\linewidth,height=4cm]{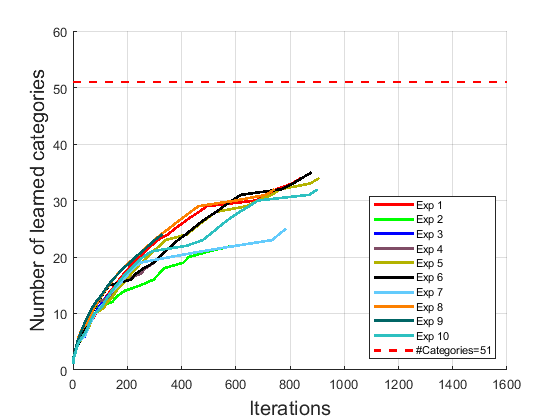}
\end{minipage}
 &
\hspace{-3mm}
 \begin{minipage}{.35\textwidth}
     \centering
     \includegraphics[width=1\linewidth,height=4cm]{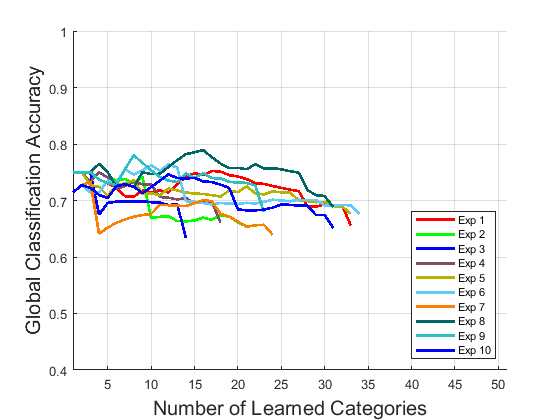}
\end{minipage}
\end{tabular}
}
\vspace{-3mm}
\\

\hspace{-8mm}
\subfloat[Summary of experiments for Local-LDA (Online Variational Inference)\vspace{-1mm}]{
\setlength\tabcolsep{0.1pt}
\begin{tabular}{c c c}

\setlength\tabcolsep{0.5pt}
\fontsize{8}{9}\selectfont
\begin{tabular}{ | c | c |c | c |c |}
\hline
 Exp\# & \#QCI & \#LC & AIC & GCA(\%)\\
\hline
1& 1346	&  40	&	12.93	&	70.51  \\ 

2& 1764	&	40	&	17.73	&	66.61  \\ 

3& 1385	&	43	&	12.4	&	70.83 \\ 

4& 1224	&	41	&	11.29	&	\textbf{72.22}  \\ 

5& 1594	&	\textbf{47}	&	13.11	&	70.20  \\ 

6& 1551	&	46	&	13.04 &	70.21  \\ 

7& 1263	&	35	&	14.83	&	67.22  \\ 

8& 1455	&	46	&	12.04	&	71.41 \\ 

9 & 1012 &	34	&	12.53	&	67.98  \\ 

10 & 1518 &		34	&	17.62	&	67.26  \\ 
\hline
\textbf{Avg.} & 1411 & \textbf{40.6} & \textbf{13.75} & \textbf{69.44}  \\ 
\hline

\end{tabular} &
\hspace{-1mm}
 \begin{minipage}{.35\textwidth}
     \centering
     \includegraphics[width=1\linewidth,height=4.3cm]{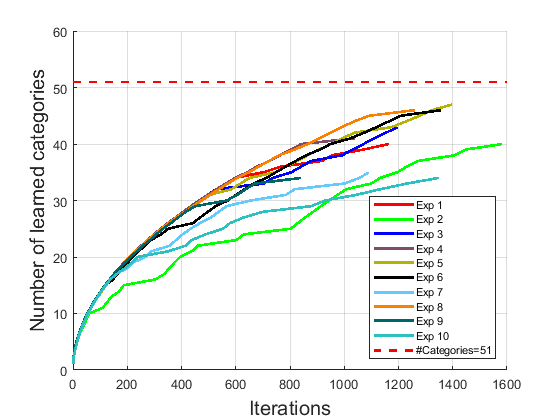}
\end{minipage}
 & 
\hspace{-4mm}
 \begin{minipage}{.35\textwidth}
     \centering
     \includegraphics[width=1\linewidth,height=4.3cm]{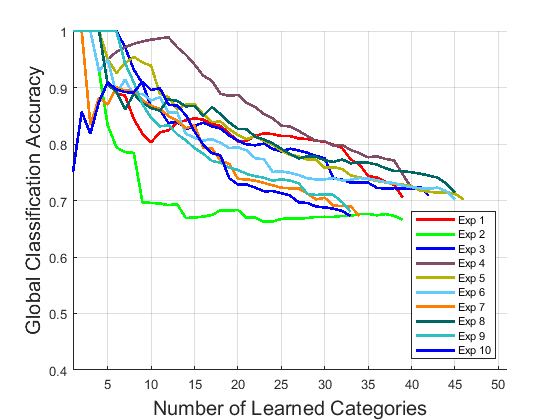}
\end{minipage}
\end{tabular}
}
 \vspace{-3mm}
 \\
\hspace{-7mm}
\subfloat[Summary of experiments for Local-HDP (our approach)\vspace{-1mm}]{

\setlength\tabcolsep{0.1pt}
\begin{tabular}{c c c}
\centering
\setlength\tabcolsep{0.5pt}
\fontsize{8}{9}\selectfont
\begin{tabular}{ | c | c |c | c |c |}
\hline
 Exp\# & \#QCI & \#LC & AIC & GCA(\%)\\
\hline
1& 1325 & \textbf{51} & 6.45 & 86.72  \\ 

2& 1370 & \textbf{51} & 8.25 & 80.44  \\ 

3& 1325 & \textbf{51} & 6.62 & 86.04 \\ 

4& 1325 & \textbf{51} & 6.70 & 85.74  \\ 

5& 1325 & \textbf{51} & 6.37 & \textbf{87.02}  \\ 

6& 1325 & \textbf{51} & 7.03 & 84.45  \\ 

7& 1325 & \textbf{51} & 6.64&	85.96  \\ 

8& 1325	&	\textbf{51}	&	6.80	&	85.36 \\ 

9 & 1330 &	\textbf{51}	&	7.17	&	83.98  \\ 

10 & 1327&	\textbf{51}	&	6.47	&	86.66  \\ 
\hline
\textbf{Avg.} & 1330 & \textbf{51} & \textbf{6.85} & \textbf{85.23}  \\ 
\hline

\end{tabular} & 
\hspace{+1mm}
 \begin{minipage}{.35\textwidth}
     \centering
     \includegraphics[width=1\linewidth,height=4.3cm]{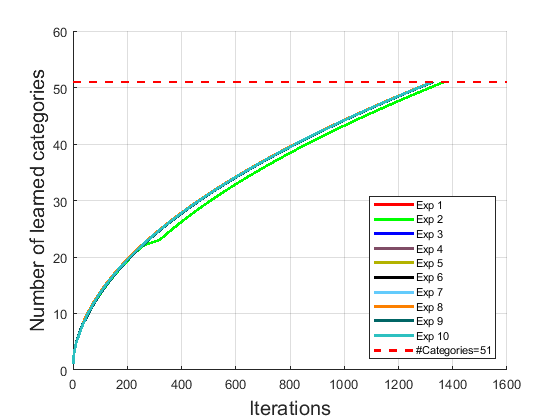}
     
\end{minipage}
 &  
\hspace{-3mm}
 \begin{minipage}{.35\textwidth}
     \centering
     \includegraphics[width=1\linewidth,height=4.3cm]{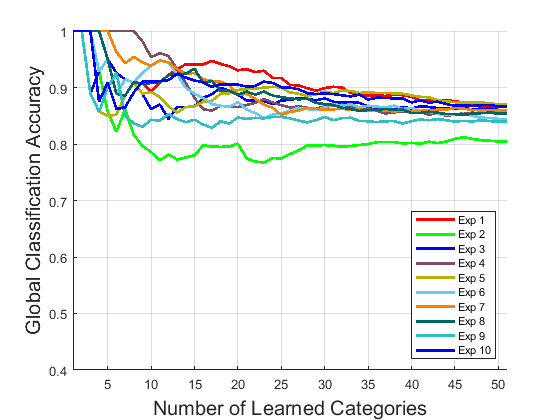}
\end{minipage}
\end{tabular}
}
\end{tabular}
\caption{Summary of 10 experiments for open-ended evaluation LDA, HDP, Local-LDA and our proposed Local-HDP approach. The learning capacity and the global accuracy of different models is compared with the corresponding plots. \vspace{-5mm}  }
\label{figOpenEndedAllExperiments}
\end{center}
\end{figure*}

\setlength{\columnsep}{15pt}%
\setlength{\intextsep}{5pt}%
\begin{wraptable}{r}{5.0cm}
\centering
\newcolumntype{?}{!{\vrule width 0.5pt}}
\setlength\arrayrulewidth{0.8pt}
\setlength\tabcolsep{1pt} 
\renewcommand{\arraystretch}{1}
\fontsize{8.5}{9.5}\selectfont
\begin{tabular}{ | c | c |c | c |c |}
\hline
 Approach & \#QCI & \#LC & AIC & GCA(\%)\\
\hline
LDA& 269 &9.1 & 16.74 & 51.00\%  \\ 

HDP& 753 &27.2 & 12.76 & 66.14\%  \\ 

Local-LDA & \textbf{1411}	&	40.6	&	13.75	&	69.44\% \\ 

Local-HDP& 1330	&	\textbf{51.0}	&	\textbf{6.85}	&	\textbf{85.23}\%  \\ 

\hline

\end{tabular}
\caption{The average result of 10 open-ended experiments for all the methods. }
\label{tbl3}
\end{wraptable}

In order to compare methods fairly, the simulated teacher shuffles data at the beginning of each round of experiments and uses the same order of object categories and instances for training and testing all the methods. Figure \ref{figOpenEndedAllExperiments}~(\textit{left}) shows the detailed summary of 10 experiments for Local-LDA, and Local-HDP methods. It shows that Local-HDP could learn all 51 categories in all experiments, while Local-LDA, HDP, and LDA, on average learned $40.6$, $27.2$, and $9.1$ categories, respectively (Table \ref{tbl3}). This result shows the descriptive power of Local-HDP.

Figure \ref{figOpenEndedAllExperiments}~(\textit{center}) shows the learning capability of the new categories as a function of the number of learned categories versus the question/correction iterations. Local-HDP achieved best performance by learning all the 51 categories in $1330.20\pm13.95$ iterations (Table \ref{tbl3}). One important observation is that shuffling the order of introducing categories by the simulated user does not have a serious effect on the performance of Local-HDP, while it affects the performance of other methods significantly. The longest experiment, on average was continued for $1411.20\pm212.75$ iterations with Local-LDA and the agent was able to learn $40.60\pm4.98$. 

Figure \ref{figOpenEndedAllExperiments}~(\textit{right}) plots the global categorization accuracy versus the number of learned categories. It was observed that the agent with Local-HDP not only achieved higher accuracy than other methods in all experiments but also learned all the categories. It is worth mentioning that Local-HDP concluded prematurely due to the ``\textit{lack of data}'' condition, i.e., no more categories available in the dataset. This means that the agent with Local-HDP has the potential of learning more categories in an open-ended fashion. According to Table \ref{tbl3}, the average GCA for Local-HDP is $85.23\%$ and it is $69.44\%$, $66.14\%$ and $51.00\%$ for Local-LDA, HDP and LDA, respectively. 


\begin{figure*}[t]
    \centering
    \includegraphics[width=1\linewidth]{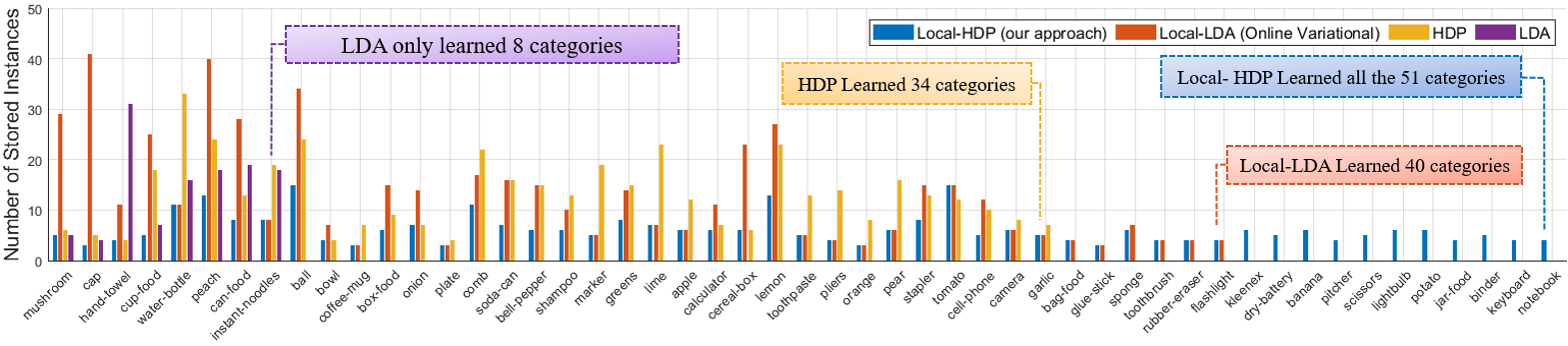}
    \caption{The absolute number of stored instances per category: the lower stored instances mean that the method is more memory efficient. The horizontal axis shows the order of introducing categories to all methods.}
    \label{fig:barplot}
\end{figure*}

Figure~\ref{fig:barplot} represents the absolute number of stored instances per category in one round of the open-ended experiments. It shows that the agent with Local-HDP stored a lower or equal number of instances for all of the categories. On closer review using Figure \ref{figOpenEndedAllExperiments}~(\textit{left}), one can see that the Local-HDP on average stored $6.85$ instances per category to learn $51$ categories, while Local-LDA stored $13.75$ to learn $40.6$ categories. HDP achieved the third place by storing $12.76$ instances to learn $27.20$ categories and LDA was the worst among the evaluated approaches, i.e., on average it stored $16.74$ instances to learn $9.10$ categories. According to this evaluation, Local-HDP is competent for robotic applications with strict limits on the computation time and memory requirements.

\begin{figure*}[h!]
\centering
\begin{tabular}{ c  c }
    \includegraphics[width=0.45\linewidth]{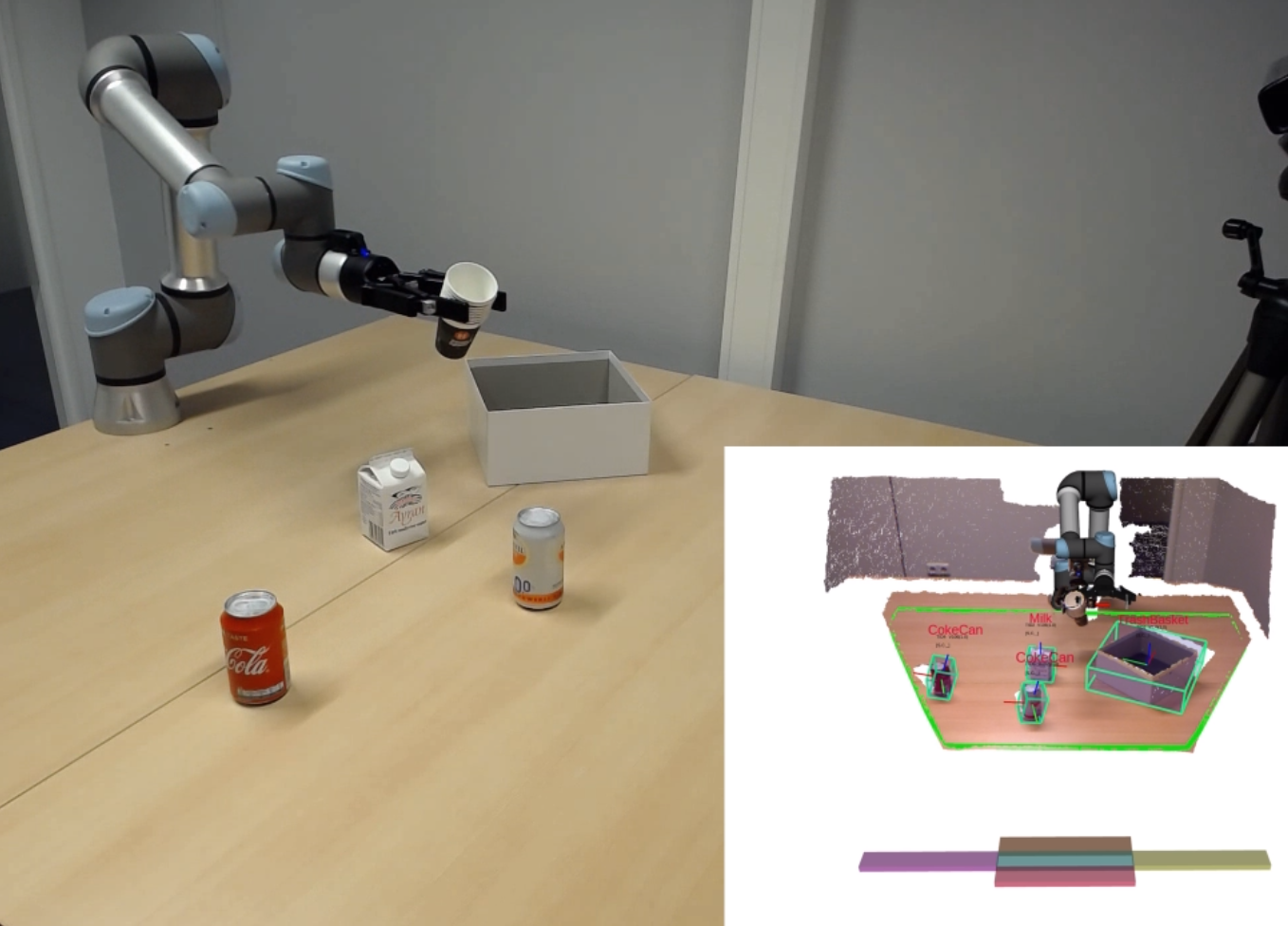} &  \includegraphics[width=0.5\linewidth]{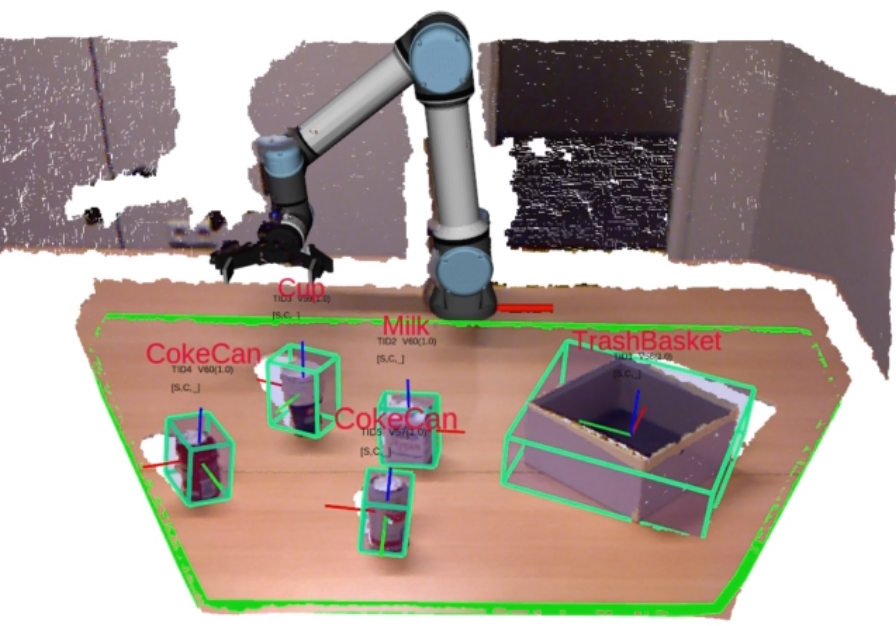} \\
    a) The robotic setup for first demonstration. &     \begin{tabular}{@{}c@{}}b) Point cloud and object category visualization in RViz \\ for the first robotic demonstration.\end{tabular}  \\
    \includegraphics[width=0.45\linewidth]{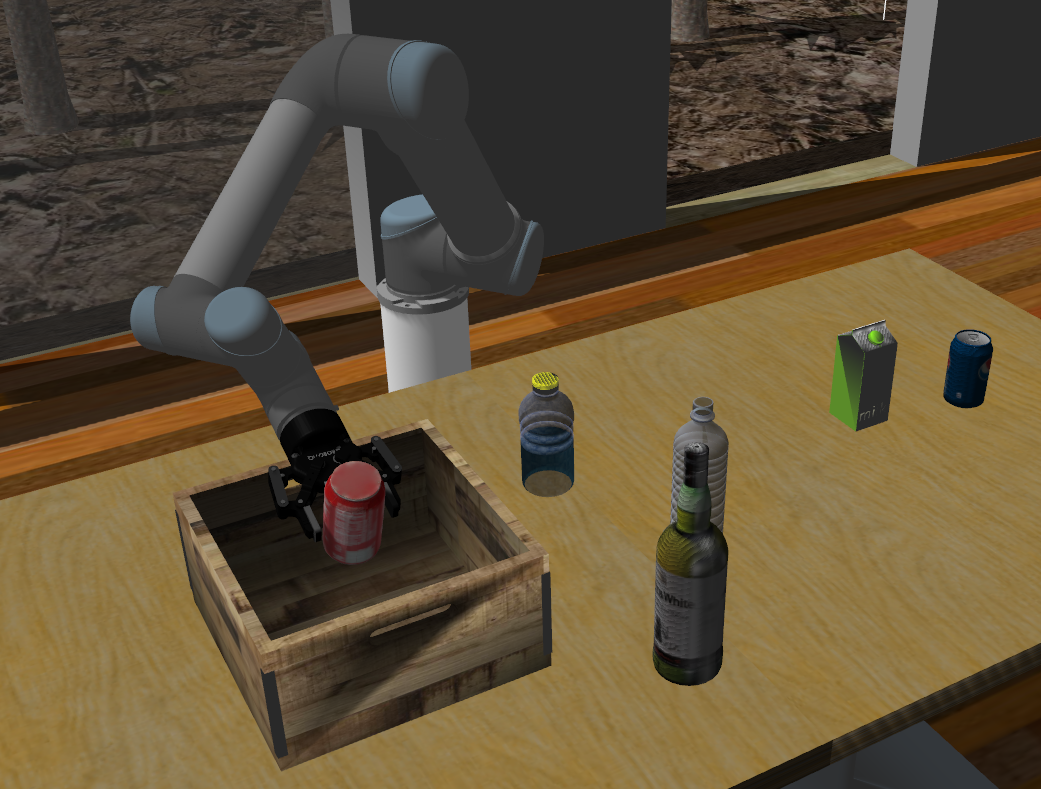} & \includegraphics[width=0.5\linewidth]{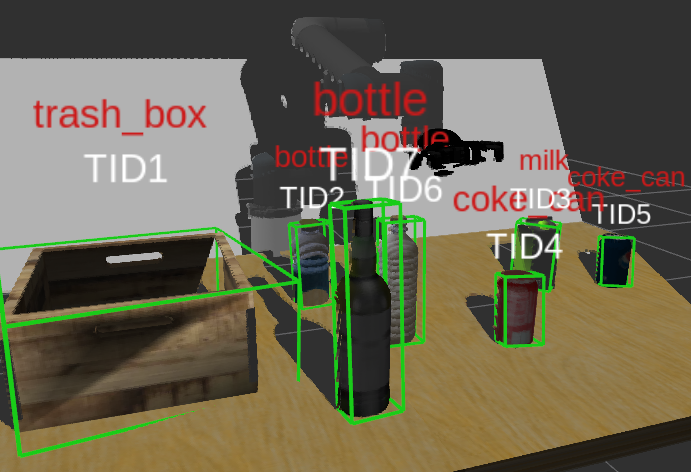}\\
    \begin{tabular}{@{}c@{}} c) Clearing coke cans from the table for the second \\ robotic demonstration.\end{tabular}
  &
    \begin{tabular}{@{}c@{}}d) The RViz visualization of the recognized categories \\ for the second robotic demonstration.\end{tabular}
\end{tabular}
\caption{The real-time application of the proposed Local-HDP 3D object category recognition method in a robotic scenario.}
\label{fig_robotic_scenario}
\end{figure*}

\section{Real-time Robotic Application}
\label{robotic scneario}
To demonstrate the applicability of the proposed 3D object categorization method in real-time robotic applications, we have performed two object-manipulation experiments, as shown in Figure \ref{fig_robotic_scenario}.

In both demonstrations, a UR5e robotic arm is used to manipulate the objects located on a table. Moreover, a Kinect camera is fixed in front of the table to acquire the visual data for further perceptual analysis. The system detects table-top objects, draws a bounding box around them and assigns a tracking ID (TID) to each object (Figures \ref{fig_robotic_scenario}.b - \ref{fig_robotic_scenario}.d). The model does not initially have any knowledge about the category of the objects located on the table. In both scenarios, we involved a human user in the learning loop as it is necessary for a human-robot interaction. In the first scenario, a user can interact with the system through the RViz\footnote{\;ROS Visualization: \url{http://wiki.ros.org/rviz}} \cite{quigley2009ros}
3D visualization environment and assign a category label to each of the detected objects on the table. After introducing the object category labels to the model, it can detect the category of the objects even if they have been placed in a different location on the table, which might change the object view partially due to the perspective or occlusion by the other objects. Finally, the clearing task is initiated in which for each individual object, the end-effector of the robotic arm moves to the pre-grasp position of a target object, and then grasps the object and put it into a trash box located on the table (Figure \ref{fig_robotic_scenario}.a). 
This demonstration showed that the system was able to detect different object categories and learned about new object categories using very few examples on-site. Furthermore, it was observed that the proposed approach was able to distinguish geometrically very similar objects from each other (e.g., \textit{Cup} vs \textit{CokeCan}). The video of this robotic demonstration is available at:  \url{https://youtu.be/YPsrBpqXWU4}

The second robotic demonstration has more emphasis on category recognition of unforeseen objects and performing a category-specific robotic task. In this demonstration, a user interacts with the system through voice commands and introduces the initially located objects on the table to the model. The model uses the segmented point cloud of these table-top objects to train the model. Subsequently, three new objects will be spawned on the table in the Gazebo simulator \cite{koenig2004design}. After the detection of each of the new objects, the system tells the predicted category to the user and asks for corrective feedback in case of a wrong prediction. This way the system learns about new object category incrementally and update a category model once a misclassification happened. 

After recognizing all object categories, the user commands the robot to clear all the coke cans from the table and put them into the trash box located on the table. To accomplish this task, the robot should detect the pose as well as the label of all objects. Then, the robot grasps and manipulates all the coke cans from the table while keeping the rest of the objects from different categories on the table (Figure \ref{fig_robotic_scenario}.c). A video for this robotic demonstration is available at: 
\url{https://youtu.be/otxd8D8yYLc}
\section{Conclusion}
\label{conclusion}
We propose a non-parametric hierarchical Bayesian model called Local Hierarchical Dirichlet Process (Local-HDP) for interactive open-ended 3D object category learning and recognition. Each object is initially represented as a bag of visual words and then transformed into a high-level conceptual topics representation. 

We have conducted an extensive set of experiments in both offline and open-ended scenarios to validate our approach and compare its performance with state-of-the-art methods. For the offline evaluations, we mainly used 10-fold cross-validation (train-then-test). Local-HDP outperformed the selected state-of-the-art (i.e., RACE, BoW, LDA, Local-LDA, and HDP) by a large margin, achieving appropriate computation time and object recognition accuracy. 
In the case of open-ended evaluation, we have developed a simulated teacher to assess the performance of all approaches using a recently proposed test-then-train protocol. Results show that the overall performance of Local-HDP is better than the best results obtained with the other state-of-the-art approaches.

 Local-HDP can autonomously determine the number of topics, even though finding a good choice for the number of topics is not a trivial task in LDA-based approaches. Moreover, the number of topics in Local-LDA should be defined in advance and is the same for all object categories, which may lead to overfitting or underfitting of the model. Local-HDP has resolved this issue by finding the number of topics for each category based on the intra-category variation of objects. Adapting online variational inference to the proposed approach enables Local-HDP to approximate the posterior for large datasets rapidly. 

In order to demonstrate the applicability of the proposed approach in real-time robotic applications, two robotic demonstrations have been conducted using a UR5e robotic arm. These experiments showed that the robot was able to learn new object categories using very few examples over time by interacting with non-expert human users.

 In the continuation of this work,  we would like to investigate the possibility of using the proposed method for graspable part segmentation of 3D objects. This way, we can address the problem of 3D object recognition and affordance detection (i.e., detecting graspable parts) simultaneously. 

\bibliographystyle{unsrt}
\bibliography{main}

\end{document}